\documentclass[acmtog]{acmart}

\newcommand{\cww}[1]{\textcolor{black}{#1}}
\newcommand{\qs}[1]{\textcolor{black}{#1}}

\copyrightyear{2025}
\acmYear{2025}
\setcopyright{acmlicensed}\acmConference[SA Conference Papers '25]{SIGGRAPH Asia 2025 Conference Papers}{December 15--18, 2025}{Hong Kong, Hong Kong}
\acmBooktitle{SIGGRAPH Asia 2025 Conference Papers (SA Conference Papers '25), December 15--18, 2025, Hong Kong, Hong Kong}
\acmDOI{10.1145/3757377.3763916}
\acmISBN{979-8-4007-2137-3/2025/12}

\acmSubmissionID{1745}

\usepackage{enumitem}
\usepackage{xspace}
\usepackage{booktabs}
\usepackage{colortbl}
\usepackage{multirow}
\usepackage{makecell}
\usepackage{lipsum} 
\usepackage{gensymb} 

\usepackage[labelsep=period]{caption}
\definecolor{MyDarkRed}{rgb}{0.46, 0.16, 0.16}
\definecolor{MyDarkBlue}{rgb}{0.16, 0.16, 0.66}

\begin{document}

\newcommand{\method}{Animus3D}

\title[\method]{\method: Text-driven 3D Animation via Motion Score Distillation}

\author{Qi Sun}
\email{qisun.new@gmail.com}
\affiliation{%
  \institution{City University of Hong Kong}
  \country{Hong Kong}
}

\author{Can Wang}
\email{cwang355-c@my.cityu.edu.hk}
\affiliation{%
  \institution{City University of Hong Kong}
  \country{Hong Kong}
}

\author{Jiaxiang Shang}
\affiliation{%
  \institution{Central Media Technology Institute, Huawei}
  \country{China}
}

\author{Wensen Feng}
\affiliation{%
  \institution{Central Media Technology Institute, Huawei}
  \country{China}
}

\author{Jing Liao$^*$}
\thanks{*Corresponding Author}
\email{jingliao@cityu.edu.hk}
\affiliation{%
  \institution{City University of Hong Kong}
  \country{Hong Kong}
}

\renewcommand{\shortauthors}{Sun \emph{et al}.}



\begin{teaserfigure}
  \includegraphics[width=\textwidth]{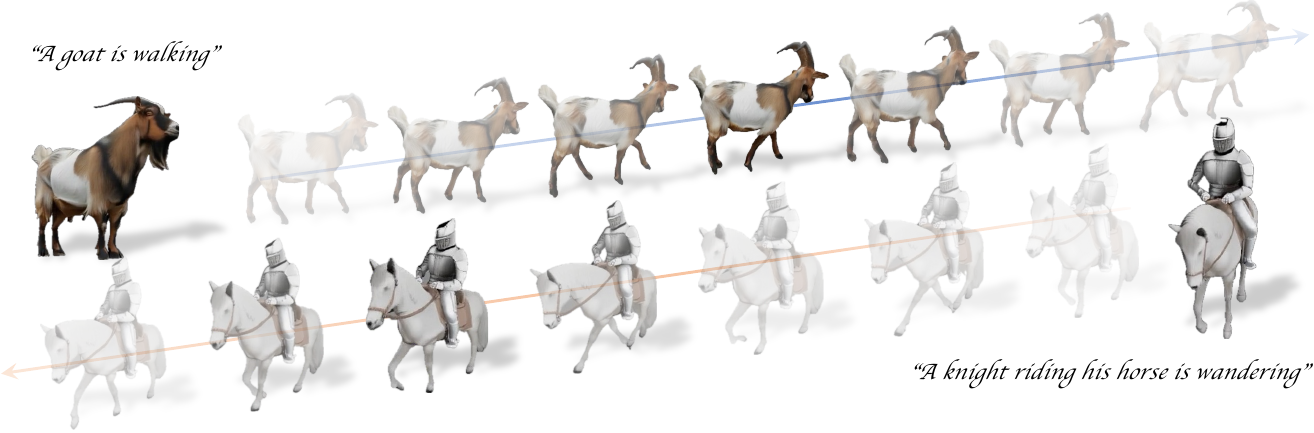}
  \caption{\textbf{Animus3D} transforms a static 3D object to an animated object sequence given text descriptions via motion score distillation.}
  \Description{This is the teaser figure for the article.}
  \label{fig:teaser}
\end{teaserfigure}

\begin{abstract}
\qs{
We present \textit{\method}, a text-driven 3D animation framework that generates motion field given a static 3D asset and text prompt.
Previous methods mostly leverage the vanilla Score Distillation Sampling (SDS) objective to distill motion from pretrained text-to-video diffusion, leading to animations with minimal movement or noticeable jitter.
To address this, our approach introduces a novel SDS alternative, Motion Score Distillation (MSD). 
Specifically, we introduce a LoRA-enhanced video diffusion model that defines a static source distribution rather than pure noise as in SDS, while another inversion-based noise estimation technique ensures appearance preservation when guiding motion.
To further improve motion fidelity, we incorporate explicit temporal and spatial regularization terms that mitigate geometric distortions across time and space.
Additionally, we propose a motion refinement module to upscale the temporal resolution and enhance fine-grained details, overcoming the fixed-resolution constraints of the underlying video model.
Extensive experiments demonstrate that \textit{\method} successfully animates static 3D assets from diverse text prompts, generating significantly more substantial and detailed motion than state-of-the-art baselines while maintaining high visual integrity.
Code will be released at \url{https://qiisun.github.io/animus3d_page}.
}
\end{abstract}

\maketitle
\section{Introduction}
\label{sec:intro}

\cww{
Text-to-3D animation has long been a foundational component of visual storytelling, entertainment, and simulation. Recent advancements demonstrate that large-scale text-to-image and text-to-video diffusion models can effectively learn valuable priors for generating 3D animations ~\cite{jiang2024animate3d, liang2024diffusion4d, sun2024eg4d, li2024vividzoo, zhang20244diffusion}.
}

\begin{figure}
    \centering
    \includegraphics[width=0.8\linewidth]{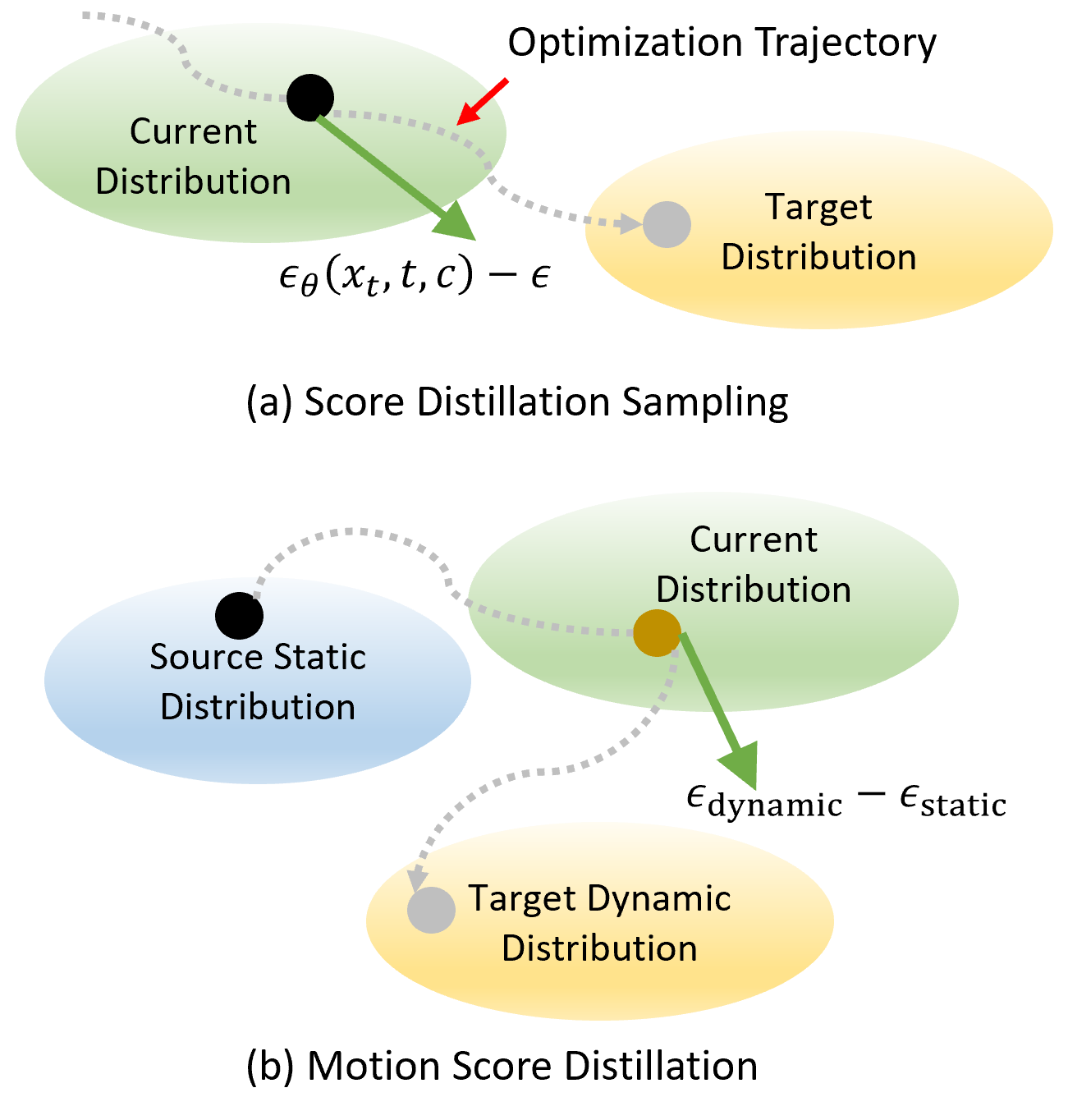}
    \vspace{-6pt}
    \caption{\textbf{Motivation.} We illustrate two distillation sampling procedures: (a) Score Distillation Sampling and our (b) Motion Score Distillation. }
    \label{fig:motivation}
    \vspace{-3mm}
\end{figure}

\cww{
A representative class of techniques that leverage such priors are score distillation sampling (SDS)-based methods~\cite{wan2025, videoworldsimulators2024, HaCohen2024LTXVideo, kong2024hunyuanvideo, hong2022cogvideo, yang2024cogvideox, he2022lvdm, xing2023dynamicrafter, chen2023videocrafter1, chen2024videocrafter2}. The core idea of SDS is to render an image of a 3D scene, add noise to the rendered image, and then use a pre-trained diffusion model to denoise it. The denoising process enables the estimation of gradients, which are then used to update the underlying 3D representation, such as neural radiance fields~\cite{mildenhall2020nerf} or Gaussian splatting~\cite{kerbl3Dgaussians}.
Recent studies have explored the theoretical foundations of Score Distillation Sampling (SDS) and formulated it as a domain transportation problem~\cite{mcallister2024rethinking,yang2023learn}, where the goal is to shift the current data distribution (i.e., rendered outputs from the 3D representation) toward a target distribution. The estimated gradient guides this transformation, as illustrated in Fig.~\ref{fig:motivation}-(a). We observe that existing SDS-based motion generation methods all adopt the original formulation of SDS without modification, inherently following this transportation framework. However, this framework reveals several key limitations for motion distillation: 1) The current distribution lacks a well-defined static source as its starting point. In motion generation, a clear static initialization is typical; the absence of its explicit modeling can obscure the starting point of the optimization trajectory, potentially resulting in limited distilled motion. 2) Motion and appearance are inherently entangled. SDS, however, does not account for this interdependency, and its estimated gradient can lead to the degradation of appearance when the distribution evolves towards the motion target.
}

\cww{
To address these challenges, we propose \textit{\method}, a text-driven 3D animation method.
At the core of our approach is a novel Motion Score Distillation strategy as depicted in Fig.~\ref{fig:motivation}-(b), which consists of two key components. First, we define a source static distribution as a canonical space, modeled using a video diffusion model enhanced with Low-Rank Adaptation (LoRA)~\cite{hu2022lora}, capable of generating static video frames. Then, we introduce a noise inversion technique. This method estimates deterministic noise for gradient computation, thereby enabling effective control of motion direction while maintaining the integrity of the appearance. We demonstrate that our Motion Score Distillation can predict more accurate transportation directions, enabling more reasonable and substantial motions while preserving the object's appearance.
Beyond the distillation process, we find motion field regularization to be crucial. To further enhance performance, we introduce temporal and spatial regularization terms into our method, which helps mitigate geometric distortions across time and space.
Additionally, due to the fixed temporal resolution of video diffusion, the motion details of animated objects are constrained. To address this, we propose an extension of our work through a motion detailization module, which extends the temporal length and enhances motion detail.
}

To conclude, we summarize our main contributions as follows:
\begin{itemize}
  \item \textbf{\textit{\method} Framework: }We propose \textit{\method}, a text-driven 3D animation framework capable of generating high-quality motion for static 3D assets from diverse text prompts.
  \item \textbf{Motion Score Distillation: }We introduce a novel Motion Score Distillation strategy that models the source static distribution using a LoRA-enhanced video diffusion model. Additionally, we {adopt} an appearance-preservation noise estimation technique, ensuring that the estimated noise during distillation does not affect the original 3D appearance.
  \item \textbf{Temporal and Spatial Regularizations:} We introduce temporal and spatial regularization terms to enhance motion field regularization, effectively reducing geometric distortions across time and space.
  \item \textbf{\qs{Motion Refinement} Module: }We address the limitation of fixed temporal resolution in video diffusion by introducing a motion refinement module, which extends the temporal length and enhances motion details for animated objects.
\end{itemize}

\section{Related Work}
\label{sec:related_works}



\subsection{Learning-based 3D Motion Generation}

\cww{
A growing body of work~\cite{liang2024diffusion4d, xie2024sv4d, ren2024l4gm, sun2024eg4d, jiang2024animate3d, zeng2024stag4d, pan2024efficient4d, wu2024cat4d} explores data-driven approaches for 3D motion generation by leveraging diffusion models to synthesize temporally consistent multi-view images, followed by pixel-wise optimization to recover coherent 3D representations with motion fields.
For instance, SV4D~\cite{xie2024sv4d} introduces temporal layers into multi-view diffusion~\cite{voleti2024sv3d}, enabling spatio-temporal modeling from monocular video inputs and supporting orbital-view synthesis of dynamic objects.
Animate3D~\cite{jiang2024animate3d} extends AnimateDiff~\cite{guoAnimateDiffAnimateYour2023c} by incorporating multi-view images, generating temporally synchronized video sequences through 3D object rendering.
Although these methods achieve impressive results on general object motion and are typically fast, they typically depend on large-scale training datasets, such as multi-view captures or densely sampled videos of dynamic scenes, which are often costly and difficult to obtain.
In contrast, our method focuses on enhancing SDS to bridge the gap between 2D generative priors and 3D animation. By distilling motion knowledge from powerful 2D diffusion models, our approach enables motion generations without requiring extensive multi-view or temporal supervision.
}


\subsection{SDS for 3D Generation}
\cww{
SDS~\cite{poole2022dreamfusion,Wang_2023_CVPR} was introduced for \qs{3D content generation and image/video editing~\cite{jeong2024dreammotion, hertz2023delta}} by distilling supervision from pre-trained 2D diffusion models~\cite{rombach2022high,ho2022imagen}.
A common issue with early SDS-based methods~\cite{poole2022dreamfusion,Wang_2023_CVPR,lin2023magic3d} is over-smoothing and lack of fine geometric or textural details. These methods often rely on high classifier-free guidance (CFG $\sim$ 100)~\cite{Classifier} to reduce output variance, which tends to cause over-saturation and unnatural results. ProlificDreamer~\cite{wang2023prolificdreamer} significantly improved fidelity by introducing a second diffusion model that is overfit to the current 3D estimate, allowing high-quality outputs with standard CFG values (e.g., 7.5). LucidDreamer~\cite{EnVision2023luciddreamer} \qs{and SDI~\cite{lukoianov2024score}} mitigates SDS's over-smoothing by replacing the random noise term with one obtained via DDIM inversion and applying multi-step denoising. 
Recent analyses such as SDS-Bridge~\cite{mcallister2024rethinking} and LODS~\cite{yang2023learn} further explore theoretical foundations and architectural optimizations for SDS-based 3D generation. Distinct from these works, our approach distills motion from a pre-trained video diffusion model by optimizing a motion field for a given static 3D object.
}

\subsection{SDS for Motion Generation}

Recent advances in video diffusion models~\cite{wan2025, videoworldsimulators2024, HaCohen2024LTXVideo, kong2024hunyuanvideo, hong2022cogvideo, yang2024cogvideox, he2022lvdm, xing2023dynamicrafter, chen2023videocrafter1, chen2024videocrafter2} have inspired a growing line of research that distills dynamic 3D scenes evolving over time from pre-trained video diffusion models. 
MAV3D~\cite{singer2023text} is one of the earliest works in text-to-dynamic object generation, introducing a hexplane representation to model scene dynamics. 
\qs{Some approaches~\cite{bah20244dfy, zheng2024unified, zhao2023animate124} uses a hybrid SDS pipeline that multi-stage optimization between supervision from text-to-image and multi-view diffusion models, improving both geometric consistency and motion fidelity.}
Other methods\qs{~\cite{li2024dreammesh4d, ling2023align, wimmer2025gaussianstolife}} explore novel 3D representations to better capture motion. AYG~\cite{ling2023align} employs 3D Gaussian Splatting~\cite{kerbl3Dgaussians, 2dgs, cui2025streetsurfgs} for efficient and high-fidelity motion representation, while Text2Life~\cite{wimmer2025gaussianstolife} introduces a training-free autoregressive approach to generate consistent video guidance across viewpoints, enhancing the quality of the distilled dynamics.
Several approaches also incorporate explicit motion priors to constrain or regularize the motion fields. TC4D~\cite{bahmani2024tc4d} uses parameterized object trajectories (e.g., translation and rotation) as motion priors.
AKD~\cite{li2025akd} further extends this idea by incorporating articulated skeletal structures into score distillation, guided by rigid-body physics simulators.
However, these methods largely adopt the original SDS formulation without modification and do not explicitly address its limitations in motion generation. In contrast, we propose a novel Motion Score Distillation strategy tailored for motion optimization, and further introduce a motion refinement module to reduce distortion caused by score distillation, resulting in more stable training and improved motion fidelity.

\begin{figure*}[tb] \centering
    \includegraphics[width=0.90\textwidth]{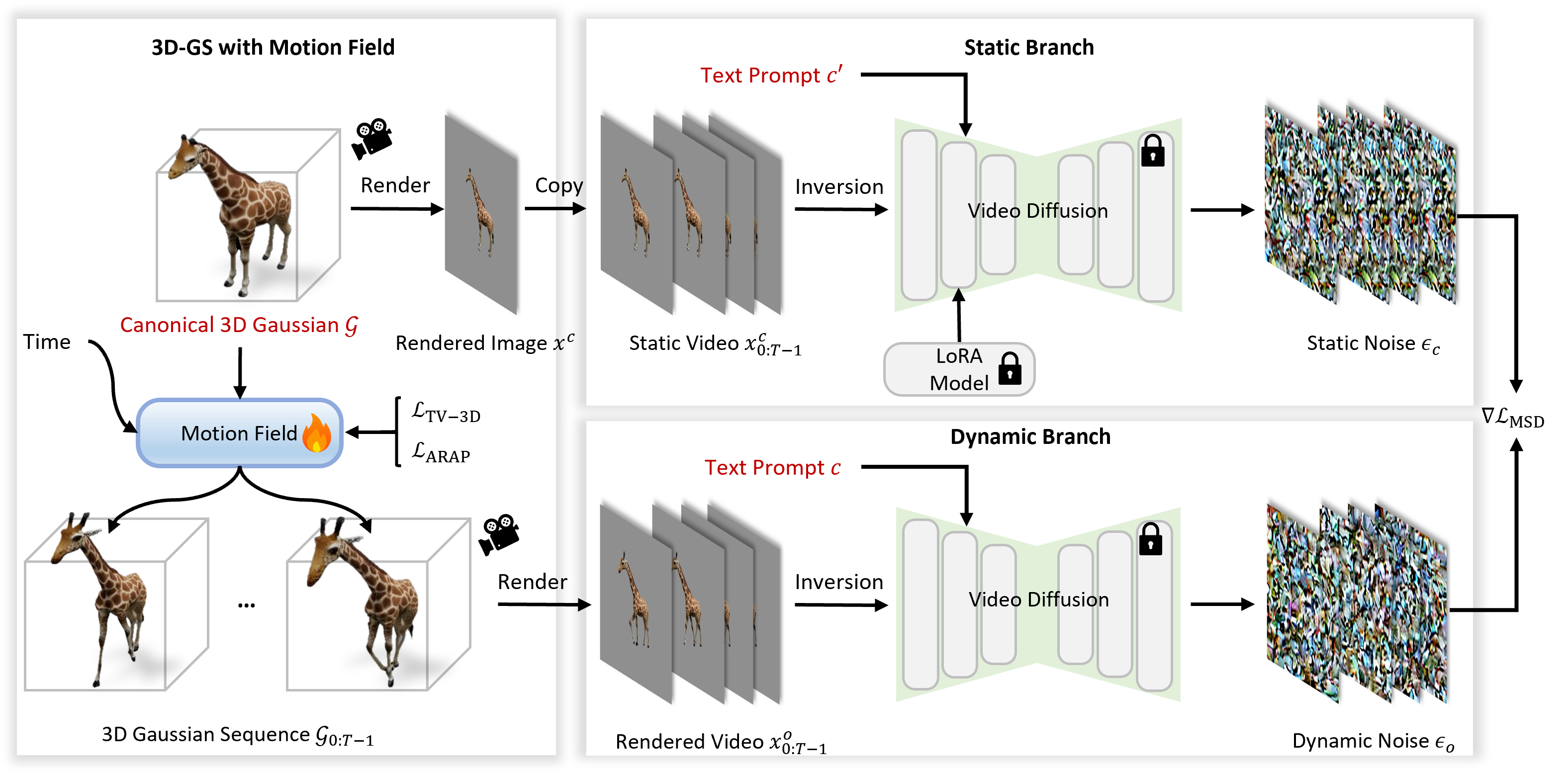}
    \vspace{-6pt}\caption{\textbf{Framework of motion generation.} 
    Given canonical 3D Gaussian $\mathcal G$, the motion field predicts the offset of Gaussian properties in each timestamp, obtaining the Gaussian sequence $\mathcal G_{0:T-1}$. 
    Then given a camera parameter, we can render the image and video from $\mathcal G$ and $\mathcal G_{0:T-1}$.
    We use video diffusion and video diffusion with LoRA to model the dynamic distribution and static distribution, respectively. 
    \cww{Given dynamic text prompt $c$ and static text prompt $c'$, the loss gradient is computed with two predicted noises.
    The gradient will guide the optimization of the motion field. We further design temporal and spatial regularization terms for the motion field to improve the performance.}} \label{fig:framework}
    \vspace{-10pt}
\end{figure*}

\section{Preliminary}
In this section, we first introduce the parametric 3D representation with motion fields. We then provide an overview of SDS.

3D Gaussian Splatting (3D-GS)~\cite{kerbl3Dgaussians} uses millions of learnable 3D Gaussians to explicitly represent a scene. Each Gaussian is defined by its center, rotation, scale, opacity, and view-dependent color encoded via spherical harmonics. The scene is rendered through a differentiable splatting-based renderer $\mathcal R_{\text{cam}}$ given camera parameters:
$x = \mathcal R_{\text{cam}}(\mathcal{G}).$
4D Gaussian Splatting (4D-GS)~\cite{Wu_2024_CVPR} extends 3D-GS by introducing a motion field to a canonical 3D representation. In our approach, we first reconstruct the static 3D object using 3D-GS, denoted as the canonical space $\mathcal{G}_c$. The motion field is modeled using a multi-resolution HexPlane with MLP-based decoders~\cite{cao2023hexplane}. During training, we keep $\mathcal{G}_c$ fixed and optimize only the motion field.
At each timestamp, the model queries the HexPlane using a 4D coordinate \qs{$(x, y, z, \tau)$} and decodes the resulting feature into deformation values for position and rotation.
By querying the motion field at each timestamp \qs{$\tau \in \{0, \dots, T-1\}$}, we generate a sequence of deformed Gaussians $\mathcal{G}_{0:T-1}$. Given camera parameters, we render the resulting $T$-frame video as
$x^o = \mathcal{R}_{\text{cam}}(\mathcal{G}_{0:T-1}).$

Score Distillation Sampling (SDS) leverages the knowledge from pretrained text-to-image diffusion models to optimize a parametric representation like 3D-GS. Given an output sample $x_0$, (e.g., a rendered image from a 3D-GS), SDS operates as follows: \textit{stochastic} Gaussian noise $\epsilon$ is added to $x_0$ at a randomly sampled timestep $t$:
\begin{equation}
    x_t = \sqrt{\bar \alpha_t}x_0 +  \sqrt{1 - \bar \alpha_t} \epsilon,
    \label{eq:addnoise}
\end{equation}
where $\bar \alpha_t$ is a noise schedule coefficient.
After that, a pretrained denoising model $\epsilon_\theta(x_t,t, c)$ predicts the noise in $x_t$, conditioned on the timestep $t$ and a text prompt $c$. SDS uses the difference between the predicted noise and the sampled stochastic noise as the gradient to update the parameterized representation:
\begin{equation}
    \nabla \mathcal{L}_{\text{SDS}} = \mathbf{E}_{t, \epsilon}[w(t)(\epsilon_\theta(x_t, t, c) - \epsilon)\frac{\partial x_t}{\partial \phi}],
\end{equation}
where $w(t)$ is the weighting function.
Recent works~\cite{mcallister2024rethinking,yang2023learn} formulate SDS as a domain transportation problem, aiming to find the optimal transport from the current data distribution $\mathcal{D}_c$ to the target distribution $\mathcal{D}_t$. Here, the rendered sample $x_0$ is drawn from $\mathcal{D}_c$ as $x_0 \sim \mathcal{D}_c$, while the text condition $c$ describes the target distribution $\mathcal{D}_t$. SDS approximates the optimal transport step $\epsilon^*$ between $\mathcal{D}_c$ and $\mathcal{D}_t$ at a given timestep $t$ by:
\begin{equation}
    \epsilon^*=\epsilon_\theta(x_t, t, c) - \epsilon
\end{equation}
Here $\epsilon_\theta(x_t, t, c)$ is a projection of
the noised image $x_t$ onto the target distribution and $\epsilon$ is a random gaussian noise $\epsilon \sim \mathcal{N}(\textbf{0},\textbf{I})$.

\section{Method}
\label{sec:method}

\cww{
Given the canonical 3D-GS $\mathcal{G}_c$ of a 3D object and a text prompt $c$ describing the desired motion, our method aims to automatically predict a motion field $f(\phi)$ for $\mathcal{G}c$. This produces a Gaussian sequence $\mathcal{G}_{0:T-1}$ that exhibits substantial, photorealistic motion while preserving the object's appearance.
To achieve this, as illustrated in Fig.\ref{fig:framework}, we first introduce Motion Score Distillation (\S\ref{subsec:msd}), an enhanced SDS framework tailored for motion learning. It incorporates dual distribution modeling~(\S\ref{subsubsec:dual}) and a appearance preservation noise estimation~(\S\ref{subsubsec:inv}) to better guide motion generation. 
We further propose temporal and spatial regularization terms~(\S\ref{sub:regularization}) to constrain the deformation fields and a Motion refinement method to extends the temporal length and enhances motion detail~(\S\ref{sub:refinement}), as shown in Fig.~\ref{fig:motion_detail}.
}



\subsection{Motion Score Distillation}
\label{subsec:msd}

\cww{
Building on the explanation of SDS in the previous section, we propose a novel approach called Motion Score Distillation (MSD). MSD aims to estimate the optimal transport from a static source distribution to a dynamic target distribution. Different from SDS, MSD approximates the optimal motion step $\epsilon^*_{\text{motion}}$ between the dual distributions at a given timestep $t$ as follows:
\begin{equation}
\epsilon^*_{motion}=\epsilon_{\text{dynamic}} - \epsilon_{\text{static}}.
\end{equation}
Thus, our MSD is formulated as follows:
\begin{equation}
    \nabla \mathcal{L}_{\text{MSD}} = \mathbf{E}_{t}[w(t) (\epsilon_{\text{dynamic}} -  \epsilon_{\text{static}}) \frac{\partial x_t}{\partial t}].
    \label{eq:main}
\end{equation}
We demonstrate that $\epsilon_{\text{dynamic}} - \epsilon_{\text{static}}$ serves as an effective gradient when both the source static and target dynamic distributions are well expressed.
Next, we detail the definitions of $\epsilon_{\text{dynamic}}$ and $\epsilon_{\text{static}}$.
}

\subsubsection{Dual distribution modeling}
\label{subsubsec:dual}

\cww{
Given a time sequence $\{0:T-1\}$, we define the static video rendered from the static 3D-GS $\mathcal{G}_c$ as $x^s_{0:T-1}$, and the dynamic video rendered from the dynamic 3D-GS $\mathcal{G}_{0:T-1}$ as $x^d_{0:T-1}$.
Similar to SDS, the target dynamic distribution can be approximated using a pretrained latent video diffusion model:
\begin{equation}
    \epsilon_{\text{dynamic}}=\epsilon_{\text{dynamic}}(x_t^d, t, c) = \epsilon_{\theta}(x_t^d, t, c).
    \label{eq:dyn_noise}
\end{equation}
Here, the text prompt $c$ describes the motion of the object, such as ``\textit{A walking <object>}''.
However, modeling the source static distribution is non-trivial. An effective static distribution should satisfy two key properties:
1) it should preserve the original semantics of the reconstructed object in the canonical space;
2) it should represent a video sequence without introducing any motion.
An intuitive solution is to model the static distribution as:
\begin{equation}
\epsilon_{\text{static}}=\epsilon_{\text{static}}(x_t^s, t, c')= \epsilon_{\theta}(x_t^s, t, c'),
\label{eq:static_text}
\end{equation}
where $c'$ is a static text description.
However, we observe that even when conditioned on a static description, the video diffusion model does not consistently generate videos of truly static objects, thus violating the second requirement.
To address this, we propose to efficiently derive a static denoiser by adapting Low-Rank Adaptation (LoRA)~\cite{hu2022lora}:
\begin{equation}
\epsilon_{\text{static}}=\epsilon_{\text{static}}(x_t^s, t, c')=   \epsilon_{\text{lora}}(x_t^s, t, c'),
\label{eq:lora_noise}
\end{equation}
where $\epsilon_{\text{lora}}$ is the lora-fined denoiser.
The LoRA parameters are trained using the static video $x^s_{0:T-1}$, with the loss function:
\begin{equation}
    \mathcal L_{\text{diff}} = \mathbf{E}_{t,\epsilon\sim \mathcal{N}(\textbf{0},\textbf{I})}[||\epsilon_\text{lora}(x_t^s, t, c') - \epsilon||^2].
\end{equation}
}

\subsubsection{Appearance preserved faithful noise estimation}
\label{subsubsec:inv}

\begin{figure}
    \centering
    \includegraphics[width=0.82\linewidth]{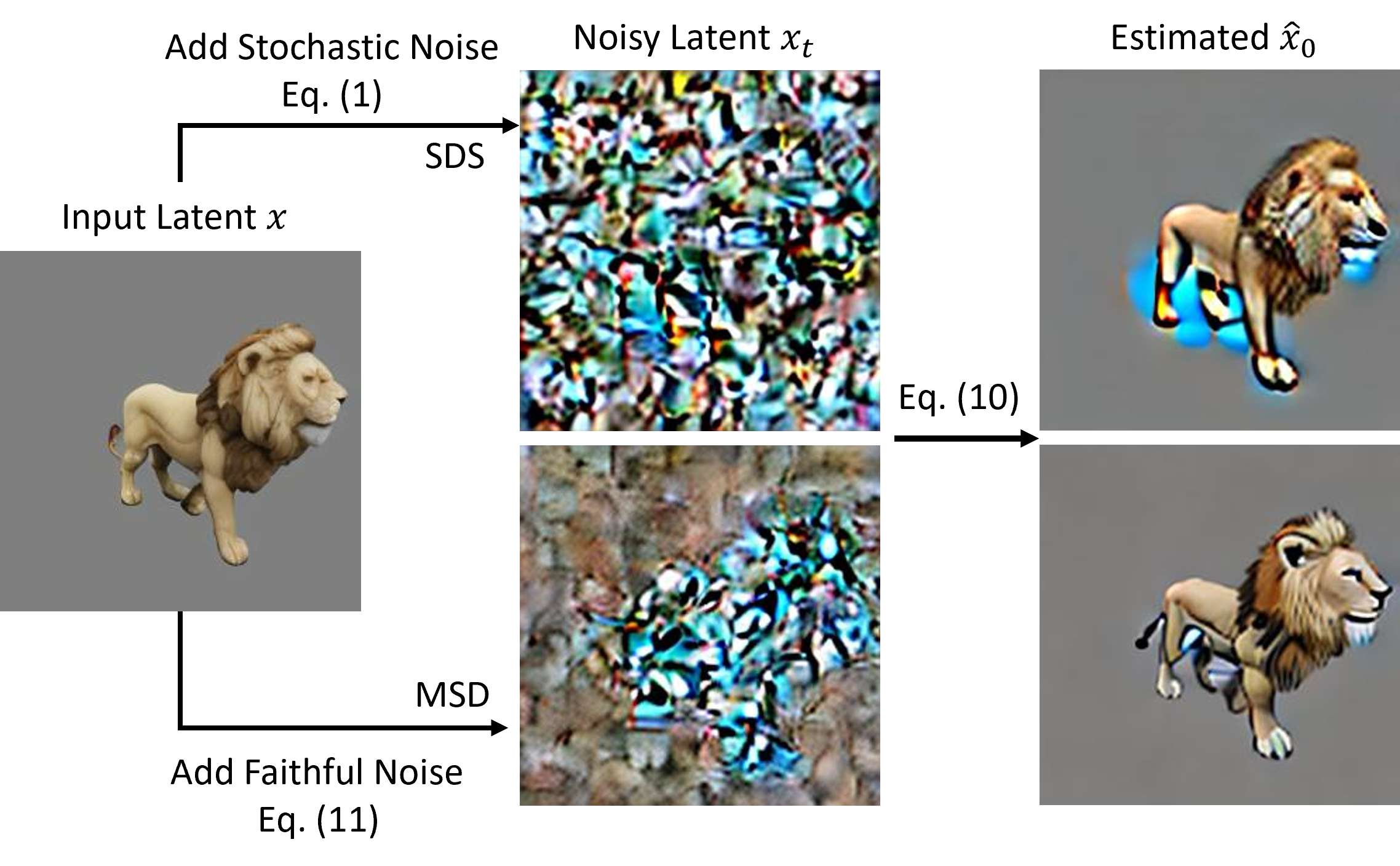}
    \vspace{-6pt}
    \caption{\textbf{Motivation of faithful noise.}
    \cww{Given an input image, we first add noise using the noise in SDS and our MSD, and then denoise it to obtain an estimated image. We find that the denoised image using SDS is significantly different from the original, exhibiting large appearance changes and background noise. In contrast, our MSD better preserves the appearance and maintains a clearer background.
    All latents are decoded into pixel space for visualization. We use t=600 in this case.}}
    \label{fig:noise-faithful}
    \vspace{-4mm}
\end{figure}

In SDS and its variants~\cite{bahmani2024tc4d,ling2023align}, we observe that it is difficult to preserve the original appearance of the static object. In some cases, the object even drifts into the background. \qs{SDS's noise estimation entangles motion and appearance, but we only optimize the motion field. This means that the appearance loss has to be compensated by geometric distortion, causing artifacts, which we refer as \textit{motion-appearance entanglement}}.
We find that this issue is strongly correlated with the stochastic noise $\epsilon$ added during the diffusion process in Eq.~\ref{eq:addnoise}.
We have assessed this observation in Fig.~\ref{fig:noise-faithful}.
Therefore, instead of adding stochastic noise, we adopt DDIM inversion~\cite{song2022denoising, lukoianov2024score} to obtain deterministic and faithful noise.
Given a noised input $x_t$, we first predict the noise $\epsilon_\theta(x_t, t, c)$ using the pretrained diffusion model.
We then estimate the corresponding denoised image $\hat{x}_0$ as:
\begin{equation}
    \hat x_0(x_t, t, c) = \frac{x_t - \sqrt{1-\bar{\alpha}_t}\epsilon_\theta(x_t, t, c)}{\sqrt{\bar{\alpha}_t}}.
    \label{eq:hatx0}
\end{equation}
Subsequently, we apply deterministic forward noising steps to obtain $x_{t+1}$ iteratively, continuing until the predefined timestep $t$:
\begin{equation}
    x_{t+1} = \sqrt{\bar{\alpha}_{t+1}} \hat x_0(x_t, t, c) + \sqrt{1-\bar{\alpha}_{t+1}} \epsilon_\theta(x_t, t, c).
\end{equation}
DDIM inversion provides a deterministic noise estimate, producing a denoised output $\hat{x}_0$ that is faithfully consistent with the input video. This facilitates appearance preservation during the optimization process.
We then apply this method to $x_t^d$ in Eq.~\ref{eq:dyn_noise} and to $x_t^s$ in Eq.~\ref{eq:lora_noise}.
\begin{equation}
\begin{split}
    x_t^d = \sqrt{\bar{\alpha}_{t}} \hat x_0(x_{t-1}^d, t, c) + \sqrt{1-\bar{\alpha}_{t}} \epsilon_\theta(x_{t-1}^d, t, c), \\
    x_t^s = \sqrt{\bar{\alpha}_{t}} \hat x_0(x_{t-1}^s, t, c') + \sqrt{1-\bar{\alpha}_{t}} \epsilon_\theta(x_{t-1}^s, t, c').
\end{split}
\end{equation}

\subsection{Motion Regularization}
\label{sub:regularization}
\cww{
To further improve performance, our method incorporates both temporal and spatial regularization terms.
}

\subsubsection{Gaussian trajectory total variation (TV-3D) for temporal regularization.}
Inspired by traditional 2D total variation (TV) losses applied in pixel space, we propose a TV-3D loss to encourage temporal smoothness in motion. This loss directly penalizes abrupt changes in the 3D positions of Gaussians across consecutive timesteps.
Specifically, it computes the $\mathcal{L}_1$ norm of the positional differences for each Gaussian between adjacent frames:
\begin{equation}
     \mathcal L_{\text{TV-3D}} = \frac{1}{N\cdot (T-1)}\sum_{\tau=1}^{T-1} \sum_{i=0}^{N} ||\mathbf x_{i,\tau} - \mathbf x_{i,\tau-1}||_1,
\end{equation}
Here, $x_{i,\tau}$ denotes the position of the $i$-th 3D Gaussian at timestep $\tau$.
By operating in the 3D Gaussian space, this constraint effectively enforces temporal consistency in the underlying geometric motion.

\subsubsection{As-rigid-as-possible (ARAP) for spatial regularization.}
To facilitate the learning of rigid motion dynamics while preserving the high-fidelity appearance of the static reference model, we employ an ARAP~\cite{arap} regularization term for spatial smoothness. 
For each gaussian point $p_j$, its ARAP contribution $\mathcal{L}_{\text{ARAP}}(p_j)$ is computed over the sequence of frames $\tau \in \{0, \cdots, T-1\}$ as:
\begin{equation}
\mathcal{L}_{\text{ARAP}} = \sum_{j=1}^N \sum_{\tau=0}^{T-1} \sum_{k \in \mathcal{N}_j} w_{jk} \left\| (p_j^\tau - p_k^\tau) - R_j^\tau (p_j^c - p_k^c) \right\|^2, 
\end{equation}
where $p_j^\tau, p_k^\tau \in \mathbb{R}^3$ are the spatial positions of point $j$ and its neighbor $k$ at the current frame $\tau$.
$p_j^c, p_k^c \in \mathbb{R}^3$ are their corresponding positions in the canonical 3D-GS.
$\mathcal{N}_j$ denotes the set of neighboring point indices for $p_j^c$, defined in the reference configuration (e.g., points within a fixed radius of $p_j^c$ ).
$R_j^\tau \in SO(3)$ is the optimal local rigid rotation for point $p_j$ at frame $\tau$. 
This ARAP loss enforces spatial consistency by encouraging locally rigid deformations, thereby promoting realistic motion while preserving geometric fidelity.

\begin{figure}[tb] \centering
    \includegraphics[width=\linewidth]{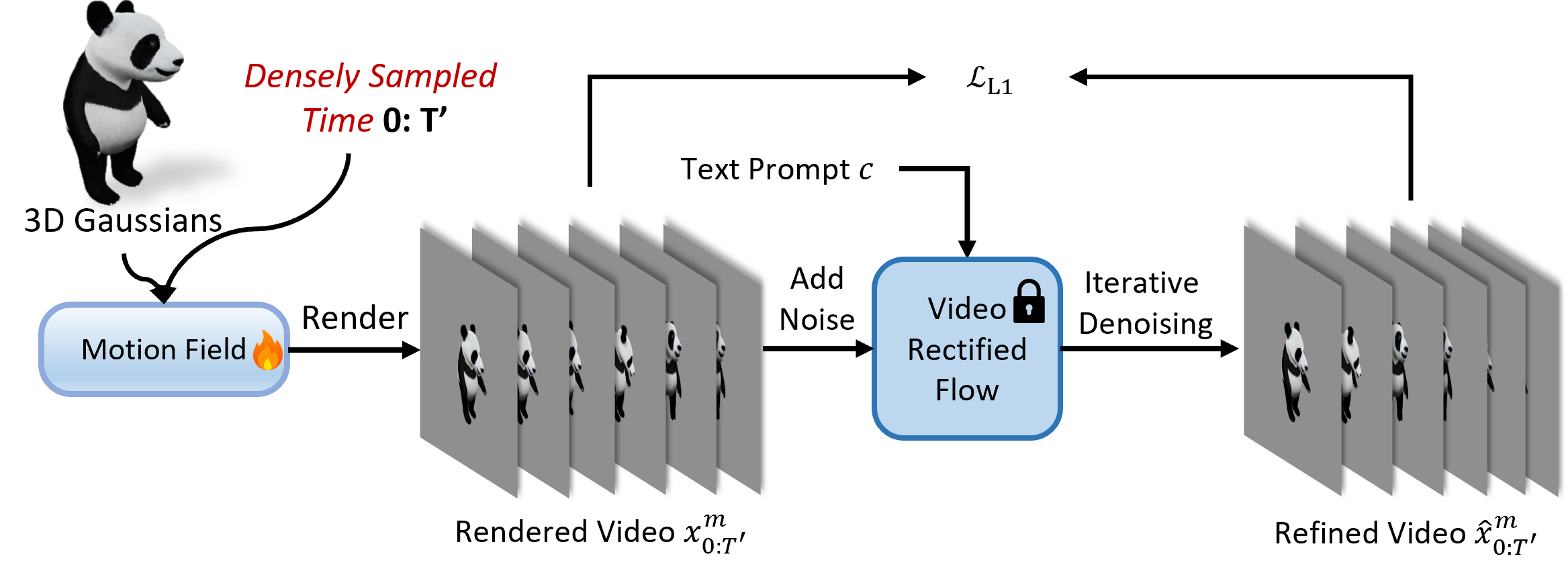}
    \vspace{-10pt}
    \caption{\textbf{Motion \qs{refinement}.} 
    \cww{We proposed a motion \qs{refinement} method to obtain fine-grained motion details.}
    } 
    \label{fig:motion_detail}
    \vspace{-10pt}
\end{figure}

\subsection{Motion Refinement}
\label{sub:refinement}
The fixed frame length in video diffusion models limits the ability of SDS-generated 3D animations to capture fine-grained motion details.
To address this, we introduce a motion refinement module that leverages a high-capacity, pre-trained rectified flow-based text-to-video model to generate long, detailed animation sequences, illustrated in Fig.~\ref{fig:motion_detail}.
Given a time sequence $\{0:T-1\}$, we interpolate it to produce $T'=2T - 1$ frames. We then render the 3D-GS with the motion field, resulting in a higher-resolution video $x_{0:T'-1}^m$ with spatial dimensions $H' \times W' \times T'$.
Following the SDEdit~\cite{meng2022sdedit} framework, we add noise to $x_{0:T'}^m$ and apply iterative denoising to generate a refined video $\hat{x}_{0:T'-1}^m$. This process preserves the original motion while enhancing temporal consistency and motion details.
Finally, we optimize the motion field using an $\mathcal{L}1$ loss between the refined video $\hat{x}_{0:T'-1}^m$ and the initial input $x_{0:T'-1}^m$. This results in a motion field capable of producing longer and more detailed animations than the original.

\begin{figure*}[tb]
    \centering
    \includegraphics[width=0.88\linewidth]{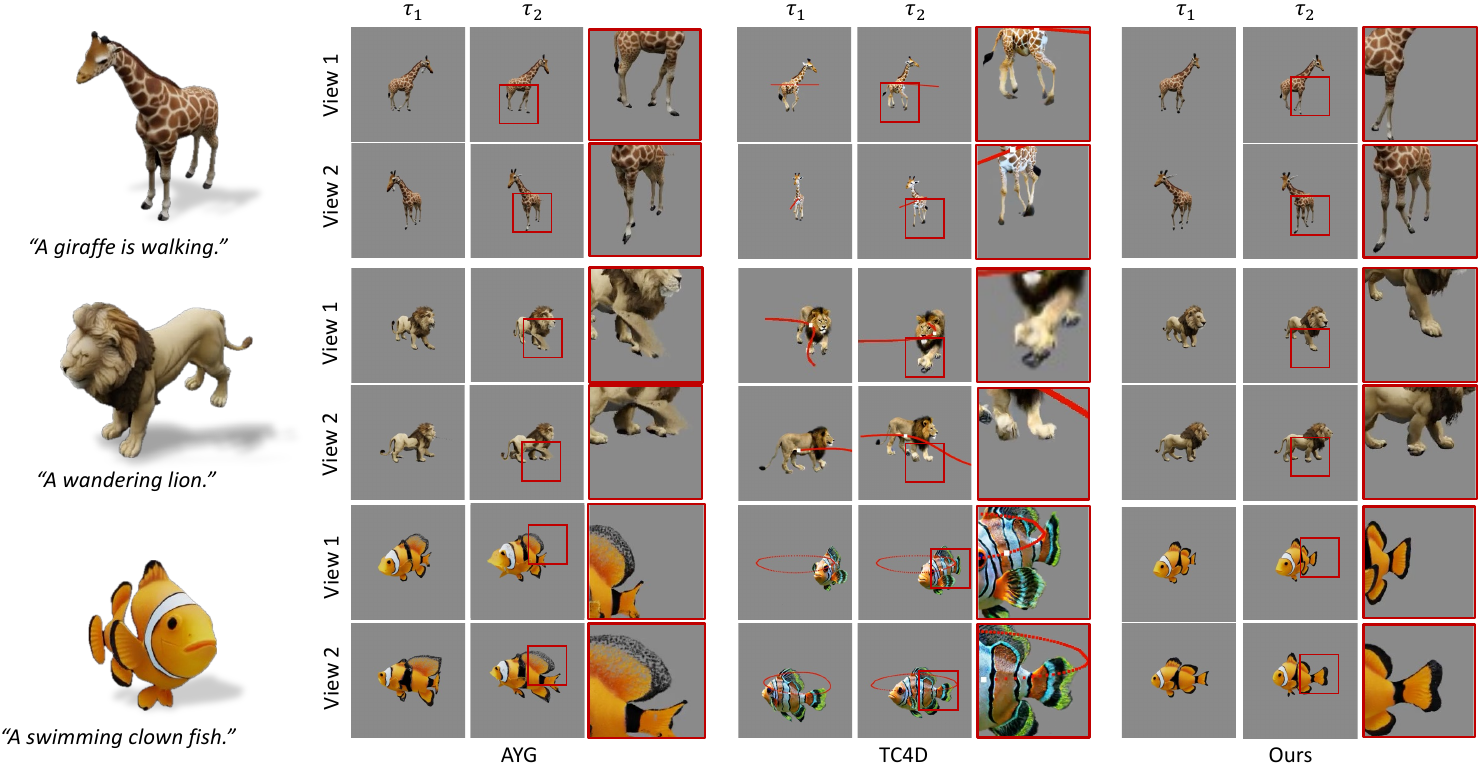}
    \vspace{-6pt}
    \caption{\textbf{Qualitative comparison with the-state-of-the-art methods.} Our methods demonstrate the substantial motion and higher visual fidelity of 3D animation. It is recommend to watch the demo for better visualization. }
    \label{fig:comparison}
    \vspace{-6pt}
\end{figure*}

\begin{figure}[tb]
    \centering
    \includegraphics[width=1\linewidth]{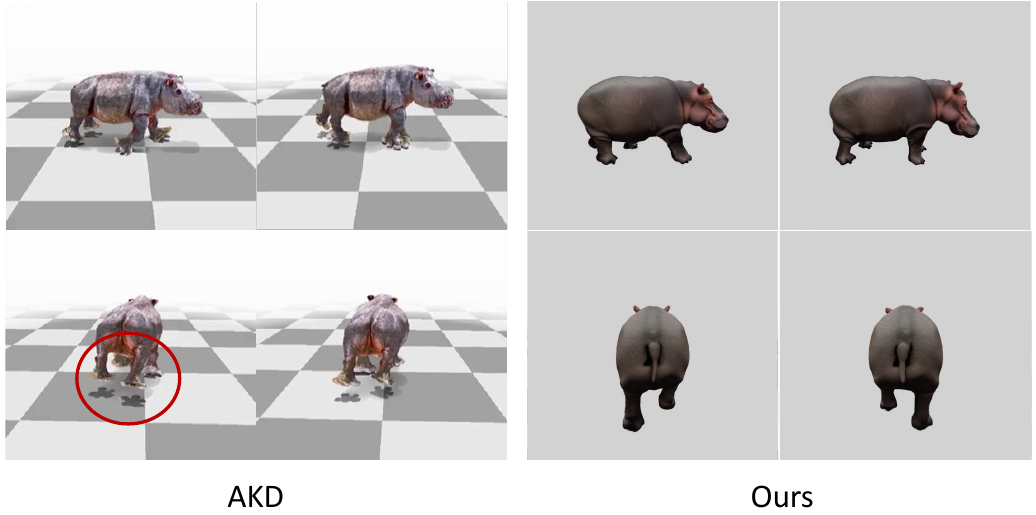}
    \caption{\textbf{Comparison with the concurrent work~\cite{li2025akd}.}
    AKD utilizes skeleton-based motion, which tends to result in perceptible stiffness in its animations. 
    }
    \label{fig:akd-comp}
    \vspace{-3pt}
\end{figure}

\begin{figure*}[htbp]
    \centering
    \includegraphics[width=0.95\linewidth]{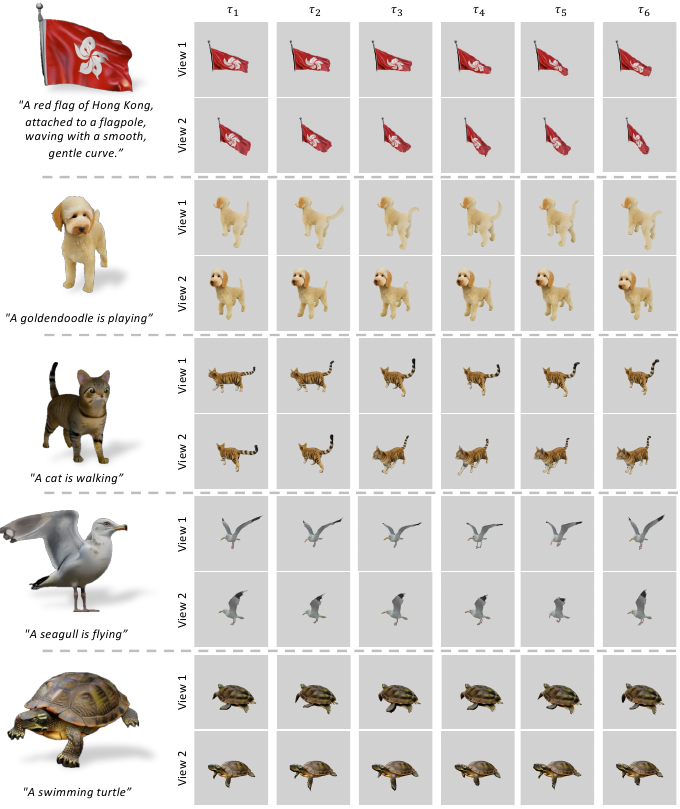}
    \caption{
    \textbf{Results of our generated 3D animations.}
     Utilizing diverse text prompts and various 3D assets, our framework generates objects faithful to their original appearance and exhibiting substantial motion. Better watch video demo to see the motion.
    }
    \label{fig:res}
\end{figure*}

\section{Experiments}
\label{sec:Experiments}

\subsection{Experimental Setup}

\paragraph{Implementation details}
\cww{
We implement our method using threestudio~\cite{threestudio2023}. 
We use ModelScopeT2V~\cite{wang2023modelscope} as the base video model. The image resolution is set to 256, with 16 frames per video.
All experiments are conducted on a single 24G GPU, with approximately 3k iterations for LoRA \qs{(rank=4, alpha=4)} fine-tuning \qs{with learning rate 1e-5}, 5k iterations for motion distillation, and 100 iterations for motion detailization.
}

\paragraph{Evaluation metrics.} 
\cww{
Following previous works~\cite{ling2023align, bahmani2024tc4d}, we utilize CLIP-image to evaluate the semantic similarity between the rendered canonical 3D-GS and the 3D-GS with the motion field. For this, we render 8 views evenly spaced around the azimuth for calculation. Additionally, we use CLIP-text to assess the alignment of the rendered object with the given text prompt. To evaluate the overall quality of the rendered videos, we compute FID~\cite{heusel2017gans} and FVD~\cite{unterthiner2018towards}.
}

\paragraph{Comparison setting.}
We compare our method with two state-of-the-art SDS-based 3D motion generation methods: AYG~\cite{ling2023align} and TC4D~\cite{bahmani2024tc4d}. For TC4D, since the authors did not release the animated 3D objects, we were unable to use the same 3D assets for animation. Instead, we extracted screenshots from their results, then used Trellis~\cite{xiang2024structured} to generate comparable 3D objects for evaluation.
Since AYG did not release their code, we opted to use the self-reproduced results for a fairer comparison. 
To ensure consistency across evaluations, we used the same 3D object generation approach as with AYG. 
\qs{We also include comparison with 4dfy~\cite{bah20244dfy} and Dream-in-4D~\cite{zheng2024unified} in the supplementary materials.}
For all comparisons, we used the same motion description. For a fair comparison, we do not employ \qs{motion refinement}~(\S\ref{sub:refinement}).

\begin{table}[tb]\centering
    \caption{\textbf{Quantitative comparison.}}
    \label{tab:comparison}
    \resizebox{0.8\linewidth}{!}{
    \large
    \begin{tabular}{lcccccc}
        \toprule
       Methods  & CLIP-Image$\uparrow$ & CLIP-Text $\uparrow$  & FID $\downarrow$ &  FVD $\downarrow$ \\
        \midrule

        AYG~\cite{ling2023align}   & 91.75 & 44.31 & 105.33 & 647.6 \\
        TC4D~\cite{bahmani2024tc4d}  & 90.99 & 50.02 & 179.14 & 340.0 \\
        Ours   & \textbf{93.04} & \textbf{51.05} & \textbf{88.50} & \textbf{204.1} \\
        \bottomrule
    \end{tabular}
    }
    \vspace{-3mm}
\end{table}

\begin{table}[tb]\centering
  \caption{\textbf{User study.}}
  \label{tab:user}
  \resizebox{0.38\textwidth}{!}{
  \large
  \begin{tabular}{lccc}
    \toprule
    {Method} & Ours  & AYG  &  TC4D \\
    Preference  & preferred (\%) & preferred (\%) &  preferred (\%) \\
    \midrule
    Overall Quality & \textbf{65.9}  & 13.4 & 20.7 \\
    Appearance Preservation & \textbf{69.5} & 11.0 & 19.5 \\
    Motion Dynamism & \textbf{75.5} & 14.2 & 10.3 \\
    Motion Text Alignment & \textbf{55.3} &  7.1 & 36.5 \\
    Motion Realism & \textbf{70.1} & 9.7 & 19.2 \\
    \bottomrule
  \end{tabular}
  }
  \vspace{-4mm}
\end{table}

\subsection{Comparisons}
\label{sub:Results I}

We report the quantitative results in Table~\ref{tab:comparison}. Our method achieves a better CLIP-Image score, as it incorporates an appearance-preserving faithful noise estimation that prevents appearance degradation during distillation. Moreover, our method outperforms others in the CLIP-Text score, demonstrating that our MSD can better generate motions that align with the user’s intension. Furthermore, the improved FID and FVD values confirm that our method produces high-quality results.
Fig.~\ref{fig:comparison} presents a visual comparison. AYG struggles to generate semantic-level and large motions due to its use of a simple SDS variant, as seen in the giraffe example. Furthermore, the motion generated by AYG is not smooth and exhibits noticeable flickering, particularly evident in the fish case.
TC4D shows simple translations along the trajectory with minimal skeleton movement. It also suffers from significant distortions and appearance changes, as demonstrated in the lion example.
While both AYG and TC4D can animate the 3D object, they fail to produce natural and realistic motions. In contrast, our framework, leveraging MSD along with regularizations, generates more substantial and realistic 3D motions.

\cww{We conduct a user preference study to evaluate performance in Table~\ref{tab:user}. 
Users are asked to evaluate five key aspects of the generated dynamic object. First, \emph{overall quality} provides a general evaluation of the rendered object. \emph{Appearance preservation} focuses on detecting any undesirable appearance deformations. \emph{Motion dynamism} assesses the extent of the object’s movement, with a preference for larger motions. \emph{Motion-text alignment} measures how well the generated motion corresponds to the text prompt. Finally, \emph{motion realism} evaluates the naturalness of the generated motion. For each aspect, users are asked to select their preferred option from AYG, TC4D, and our method. We received 17 valid responses and present the user preference rates for each aspect. Our method achieves the highest preference rate across all aspects, demonstrating that it generates more natural and realistic motions.}

\begin{figure}[tb]
    \centering
    \includegraphics[width=0.82\linewidth]{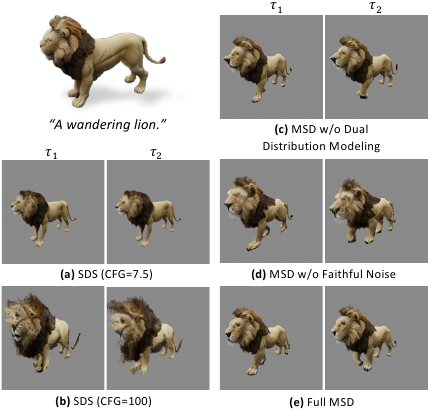}
    \caption{\textbf{Comparison between different score distillation methods.} 
    }
    \label{fig:abl-distill}
\end{figure}

\begin{figure}[tb]
    \centering
    \includegraphics[width=0.82\linewidth]{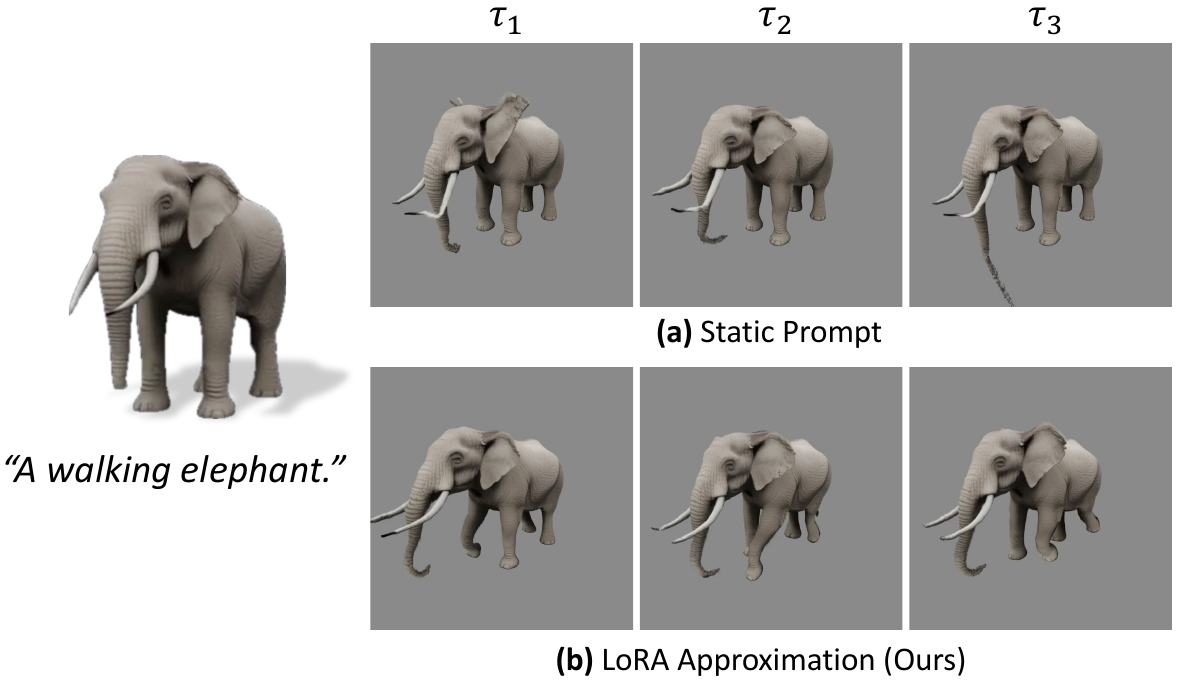}
    \caption{\textbf{Ablation on different static distribution modeling methods.} }
    \label{fig:static}
    \vspace{-6pt}
\end{figure}

\begin{figure}[tb]
    \centering
    \includegraphics[width=1\linewidth]{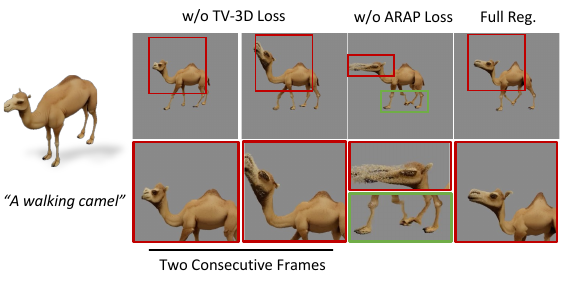}
    \caption{\textbf{Effectiveness of motion regularization.}
    Without TV-3D loss, the camel demonstrate the large object deformation in consecutive frames, while without ARAP loss, the camel would perform non-rigid motion.
    }
    \label{fig:motion-reg}
    \vspace{-6pt}
\end{figure}

\subsection{Ablation Studies}

\subsubsection{Comparison between different score distillation methods.}
\cww{We compare our motion distillation approach across several variants: (a) SDS with CFG=7.5, (b) SDS with CFG=100, (c) our method without faithful noise, (d) our method without dual distribution modeling, and (e) our full MSD approach. No motion regularization is applied in these experiments to ensure a fair comparison.
From variant (c), we observe that generating substantial and large motion is difficult without explicitly modeling the static distribution. While variant (d) produces sufficient motion, the lack of faithful noise significantly compromises the original appearance fidelity and introduces notable background artifacts.
Without our MSD, the approach degrades to conventional SDS, which either fails to generate noticeable motion with a small CFG (a) or results in meaningless motion and severe appearance distortion with a large CFG (b). In contrast, our full MSD method is uniquely effective in generating realistic motion while robustly preserving the original appearance fidelity.}

\subsubsection{Denoising with approximated static distribution.} 
\cww{As discussed in \S\ref{subsubsec:dual}, an alternative approach to defining the source static distribution is to use a static text prompt, such as ``\textit{low motion, static statue, not moving, no motion, <object>}'', as described in Eq.\ref{eq:static_text}. However, we observe that even when conditioned on such a static description, the video diffusion model does not consistently generate videos of truly static objects and often leads to the introduction of local distortions during distillation, as shown in Fig.\ref{fig:static}.
In contrast, by using a LoRA-enhanced model to model the static distribution, we can effectively induce large, reasonable motions.}

\subsubsection{The effectiveness of motion regularization.}
As illustrated in Fig.~\ref{fig:motion-reg} (a), the TV-3D loss encourages temporal smoothness in Gaussian points, and removing this loss results in large displacements between consecutive frames. In panel (b), we show that the ARAP loss provides a crucial spatial constraint for maintaining rigid motion; without this loss, the object experiences significant distortion or may even break apart.
Therefore, we incorporate both temporal and spatial regularization terms in our method to further enhance the results of MSD.

\begin{figure*}[tb]
    \centering
    \includegraphics[width=1\linewidth]{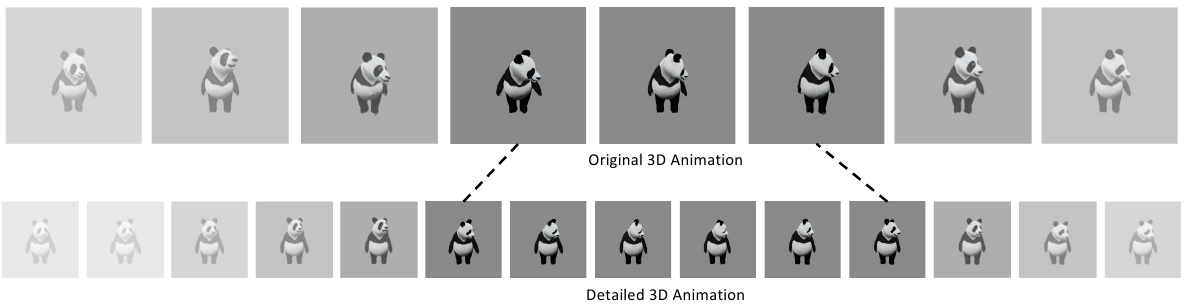}
    \caption{\textbf{Demonstration of motion refinement.} Limited by the frame length of the underlying video diffusion, SDS could only generate fixed-length animation, leading to non-continuous motion (large motion change in consecutive frames) in the original 3D animation; Our motion refinement could generate more fine-grained motion (longer 3D animation) benefited from additional larger video diffusion model.}
    \label{fig:motion-detailization}
    
\end{figure*}

\begin{figure}[tb]
    \centering
    \includegraphics[width=1\linewidth ]{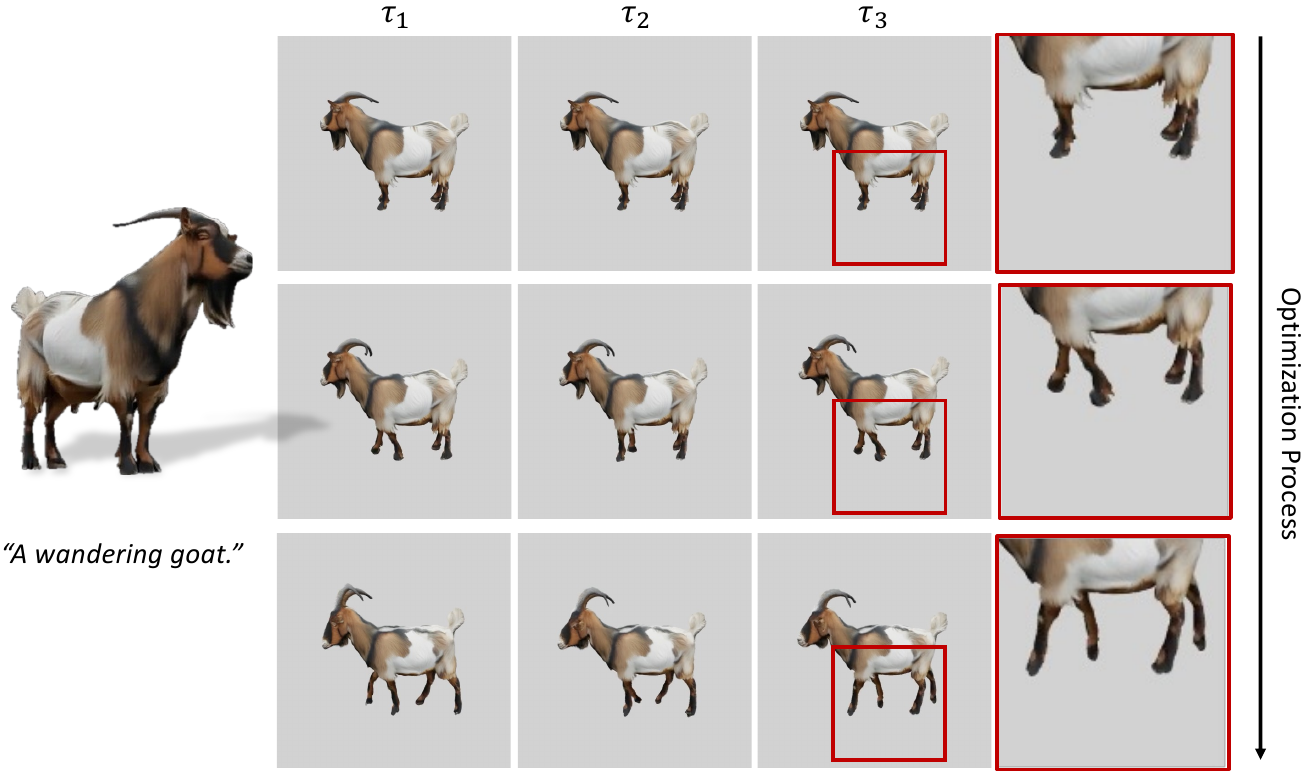}
    \caption{\textbf{Visualization of motion generation process.} With increasing training iterations, 3D object could generate more substantial motion with MSD optimization.}
    \label{fig:vis}
    \vspace{-10pt}
\end{figure}

\subsection{Visual Results}

We illustrate the motion distillation process in Fig.~\ref{fig:vis}, where our framework progressively optimizes the motion field of the 3D object (goat) over multiple optimization steps. This allows the 3D object to appear ``moving'' while preserving its original appearance.
\cww{Due to the frame length limitation of the underlying video diffusion, SDS can only generate fixed-length animations, resulting in non-continuous motion (i.e., large motion changes between consecutive frames) in the original 3D animation. In contrast, our \qs{motion refinement} approach generates more fine-grained motion (longer 3D animations) by leveraging a larger video diffusion model. As shown in Fig.~\ref{fig:motion-detailization}, our \qs{motion refinement} transforms non-continuous motion, caused by the fixed-length video diffusion, into more natural and smooth animation sequences.}
We present additional results in Fig.~\ref{fig:res}. Our method supports a variety of motion descriptions, such as "playing," "walking," "flying," and "swimming." Furthermore, our approach can animate not only animals but also general objects, as demonstrated with the red flag of Hong Kong.

\section{Conclusion}
\label{sec:Conclusion}
\cww{In this paper, we introduce Motion Score Distillation (MSD) for text-driven 3D animation. Our method formulates score distillation as distribution transportation, enhancing conventional techniques through dual distribution modeling and faithful noise. Specifically, we tackle the challenge of static video distribution modeling by using LoRA-enhanced video diffusion, and we perform appearance-preserving faithful noise estimation to mitigate the appearance changes often encountered in SDS.
Additionally, we integrate spatial-temporal geometric motion regularizations and apply motion detailization using large video models to ensure scalability. Experimental results on text-driven 3D animation, along with comprehensive ablation studies, demonstrate that our method outperforms current state-of-the-art approaches and validates its effectiveness.}

\begin{figure}[tb]
    \centering
    \includegraphics[width=\linewidth]{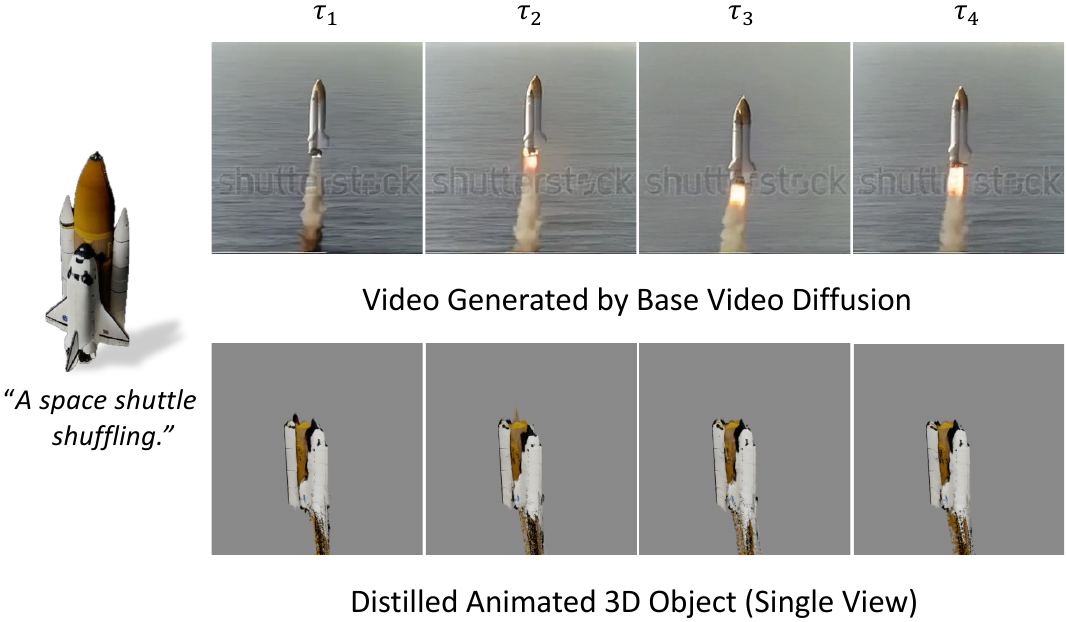}
    \caption{\textbf{Failure case}. Our  motion could only deform the given 3D object, which can hardly model fluid ejected from inside a rocket.}
    \label{fig:failure}
    \vspace{-3pt}
\end{figure}

In Fig.~\ref{fig:failure}, we illustrate a limitation of our method: it struggles to model new content appearing in the scene, such as ejected fluid. Instead of generating this new content, the model tends to introduce distortions as a form of compensation. This issue could potentially be addressed in the future by designing new particle generation and modeling strategies.
Another challenge is the optimization time—a common drawback of score distillation methods—which typically requires several hours. This inefficiency could be mitigated through techniques such as amortized training~\cite{lorraine2023att3d} or the adoption of more efficient data structures~\cite{mueller2022instant}.

\begin{acks}
We thank all anonymous reviewers and area chairs for their valuable comments. 
This work was supported by Central Media Technology Institute, Huawei [Project No. TC20240927042] and a GRF grant from the Research Grants Council (RGC) of the Hong Kong Special Administrative Region, China [Project No. CityU 11208123].
\end{acks}


\bibliographystyle{ACM-Reference-Format}
\bibliography{ref}


\appendix



\end{document}